\documentclass[10pt,journal,compsoc]{IEEEtran}
\newif\ifpeerreview

\peerreviewtrue

\usepackage[nocompress]{cite}
\usepackage{url}
\usepackage{amsmath,amssymb,graphicx}
\usepackage{algorithm}
\usepackage{algpseudocode}

\usepackage{lipsum} 

\usepackage[switch]{lineno}

\title{Object Motion Sensitivity: A Bio-inspired Solution to the Ego-motion Problem for Event-based Cameras}

\author{Shay Snyder$^\dag$, Hunter Thompson$^\dag$, Md Abdullah-Al Kaiser, \\Gregory Schwartz, Akhilesh Jaiswal, and Maryam Parsa*
\IEEEcompsocitemizethanks{
    \IEEEcompsocthanksitem \{$\dag$\} authors contributed equally\protect
    \IEEEcompsocthanksitem \{*\} corresponding author - mparsa@gmu.edu
    \IEEEcompsocthanksitem Shay Snyder and Maryam Parsa are with George Mason University\protect
    \IEEEcompsocthanksitem Hunter Thompson is with the Georgia Instute of Technology\protect
    \IEEEcompsocthanksitem Md Abdullah-Al Kaiser and Akhilesh Jaiswal are with the University of Southern California\protect
    \IEEEcompsocthanksitem Gregory Schwartz is with Northwestern University}
}

\begin{document}

\IEEEtitleabstractindextext{%
\begin{abstract}
Neuromorphic (event-based) image sensors draw inspiration from the human-retina to create an electronic device that can process visual stimuli in a way that closely resembles its biological counterpart.
These sensors process information significantly different than traditional RGB sensors. 
Specifically, sensory information generated by event-based image sensors is orders of magnitude sparser compared to that of RGB sensors.
The first generation of neuromorphic image sensors, Dynamic Vision Sensors (DVS), are inspired by the computations confined to the photoreceptors and the first retinal synapse. 
In this work, we highlight the capability of the second generation of neuromorphic image sensors, Integrated Retinal Functionality in CMOS Image Sensors (IRIS), which aims to mimic full retinal computations from photoreceptors to output of the retina (retinal ganglion cells) for targeted feature-extraction.
The feature of choice in this work is Object Motion Sensitivity (OMS) that is processed locally in the IRIS sensor.
Our results show that OMS can accomplish standard computer vision tasks with similar efficiency to conventional RGB and DVS solutions but offers drastic bandwidth reductions.
This cuts the wireless and computing power budgets and opens up vast opportunities in high-speed, robust, energy-efficient, and low-bandwidth real-time decision making.
\end{abstract}

\begin{IEEEkeywords} 
event-based vision sensors, retinal computation
\end{IEEEkeywords}
}

\maketitle

\IEEEraisesectionheading{
  \section{Introduction}\label{sec:introduction}
}
%
%
%
%
\IEEEPARstart{D}{igital} cameras have become an essential tool for capturing visual information in our environment. Their applications range from smartphones \cite{Hayes:12} and autonomous driving~\cite{ma20223d}, to robotics \cite{snyder2021thor}, and manufacturing \cite{zhou2022computer}. Event-based cameras, also referred to as neuromorphic cameras, represent the next generation of imaging technology, drawing inspiration from biological retinas to extract events from visual stimuli in a faster and more efficient manner compared to traditional cameras \cite{gallego2020event}. 
 The most common event-camera architecture is the Dynamic Vision Sensor (DVS), which is made up of a pixel array that responds asynchronously and independently to brightness changes in the scene \cite{gallego2020event}. 
 
This continuous stream of events is different from the sequential production of frames in traditional active pixel sensor (APS) cameras.
Importantly, events are sparse in space and time and therefore enable a memory and energy efficient representation of spatiotemporal activity including motion.
Therefore, DVS serves as a practical solution to the size, weight, and power (SWaP) constraints of embedded image processing systems, such as self-driving cars \cite{binas2017ddd17} and autonomous robotics \cite{binas2017ddd17}.

In real-world applications of event-based computer vision systems, distinguishing between events caused by moving objects and those caused by the camera's ego-motion has been a persistent problem \cite{khan2017ego}. Unlike RGB frames, event data provides limited contextual information about the observed scene. Consequently, it is challenging to distinguish between the active foreground object's spikes and the static background events caused by the camera's ego-motion. Numerous methods have been proposed to address the ego-motion problem, ranging from incorporating inertial data to fitting a linear motion model \cite{jia2021self, kinsman2012ego}. While some recent approaches employ neural networks to estimate ego-motion, these methods tend to be computationally intensive.

A biological solution emerges from a computation performed in the neural circuitry of the animal retina--\textit{Object Motion Sensitivity (OMS)} \cite{schwartz2021object, baccus2008retinal}.
OMS is a fundamental computation performed within the animal visual system by the feature-spike activity of Retinal Ganglion Cells (RGCs). The algorithm instantiated in the biological circuit involves subtracting a global temporal contrast signal in the receptive-field surround from a local contrast signal in the receptive-field center.

\begin{figure}
    \centering
    \includegraphics[width=0.37\textwidth]{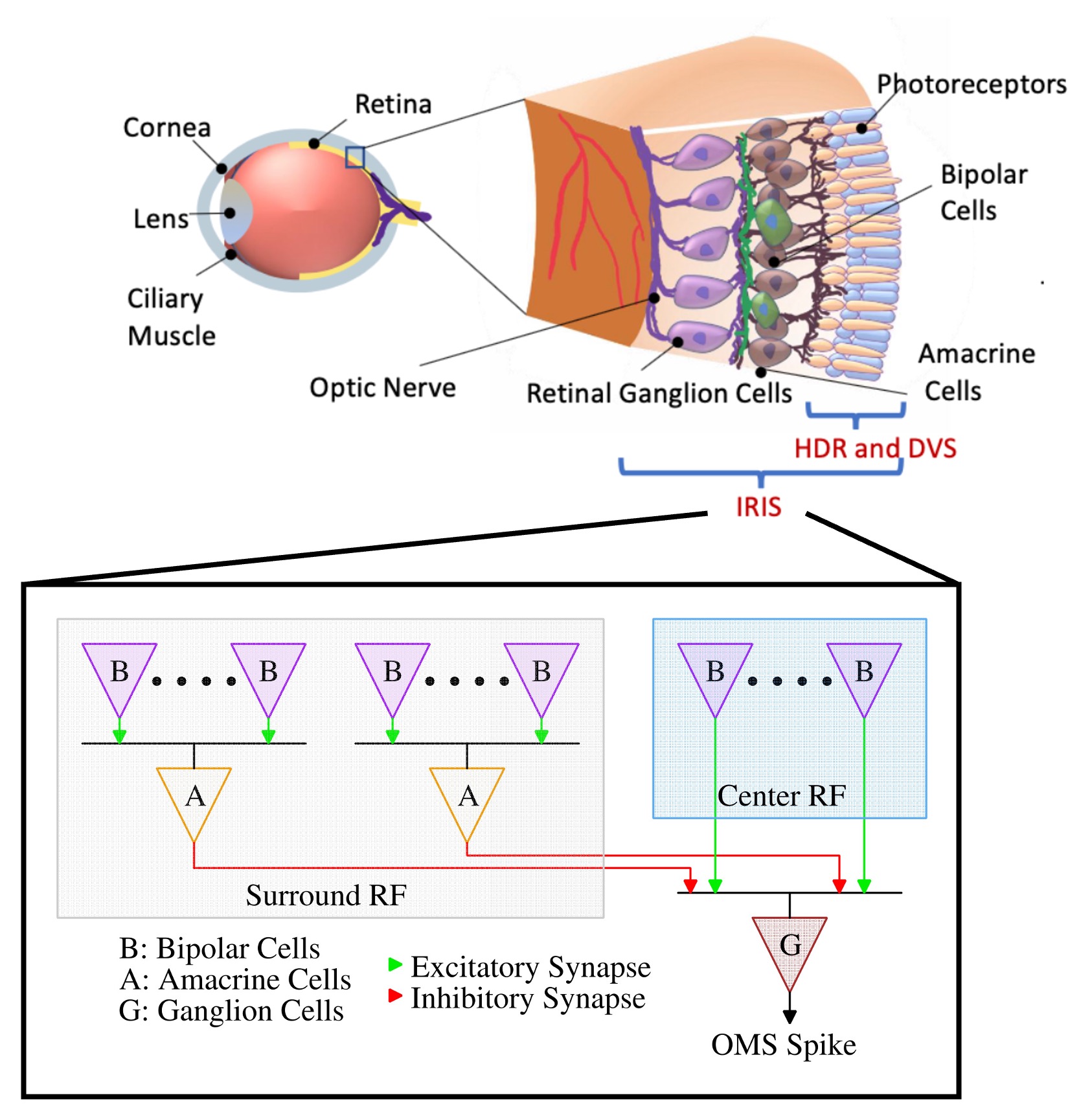}
    \caption{Retinal circuit for Object Motion Sensitivity embedded inside hierarchical retinal layers.}
    \label{fig:biological_oms}
\end{figure}

Figure \ref{fig:biological_oms} visualizes the biological retina's architecture that is responsible for extracting the OMS features and highlights that OMS aims to build upon the biological underpinnings of DVS to develop a more biologically plausible sensor.
This activity is used by the brain to discriminate the motion of objects (object motion) from motion caused by motion of the observer (ego-motion) \cite{schwartz2021object}.

This work takes inspiration from the biological retina to evaluate in-sensor computation methods in real-world environments. Integrated Retinal Functionality in CMOS Image Sensors (IRIS) is a novel retina-inspired approach to vision sensing that uses spike-based processing to filter out the dichotomy between self motion of the camera from the physical movements of objects in the scene \cite{yin2022iris}. Furthermore, IRIS reduces the computational requirements of machine learning models by incorporating data preprocessing and feature extraction within the physical sensor leveraging 3D semiconductor integration technology. This approach stands at the center of the edge computing paradigm where budgets for wireless and computing power can be drastically reduced along with opening up ground breaking opportunities in ultra-fast decision making. Moreover, novel application driven chip solutions like IRIS are at the epicenter of United States semiconductor road map (CHIPS act) \cite{Rep_Ryan_2022}. 

In summary, the major impacts of this paper are as follows:

\begin{enumerate}
    \item We present an algorithmic model inspired by biological retinal ganglion cell (RGC) computations for extracting OMS features from visual stimuli. Our evaluation focuses on the effectiveness of the model in capturing OMS features compared with RGB and DVS.

    \item We assess the performance characteristics of OMS in a standard computer vision task; object detection using a high-resolution autonomous driving dataset along with a state-of-the-art convolutional neural network.

    \item We perform a thorough evaluation of the numerous benefits that come from OMS where the resulting representation contains 3.26x more information per bit of transmitted data. 
    
\end{enumerate}

\section{Related Work}

\noindent
To the best of our knowledge, this works serves as a foundational work at the intersection of end-to-end retinal computations applied to existing computer vision tasks. As such, we go through two research areas that are directly related to this novel application: ego-motion compensation and optimization for size, weight, and power constrained environments.

\textbf{Ego-Motion Compensation} The ego-motion problem is a persistent issue which has plagued efforts to use DVS cameras mounted on moving platforms. Whenever the platform shifts, the DVS pixels pick up on the reflectance changes and produce output even when all entities in the scene remain static. These events tend to occur around edges in the scene but frequently extend to the object surfaces. This results in the occlusion of salient entities in the scene which complicates the task of moving object detection. Many works have addressed this problem through the parametric modeling of camera motion.

Stoffregen, et al. \cite{stoffregen2019event} leverages an iterative Contrast-Maximization \cite{gallego2018unifying} approach to jointly model the motion parameters by defining a set of clusters and predicting their event-cluster assignments. While this approach is effective, it requires a slow and computationaly intensive iterative approach.
A later work by Liu, et. al. \cite{liu2020globally} attempts to solve this problem through the use of bounding functions to place constraints on Contrast-Maximization though they still required the use of gradient descent and only evaluated their work on rotational ego-motion.

Similarly, Mitrokhin, et al. \cite{DBLP:journals/corr/abs-1803-04523} introduces an approach to object tracking compatible with event-based cameras using parametric models. These modes are used to estimate the three-dimensional geometry of the event data and correct for ego-motion noise. The central experiment within this paper is the ability of the model to segment object motion and compensate for ego motion. Compared to OMS and IRIS, this work has two major limitations: (1) there are no references to biological inspirations, and (2) it cannot be embedded into low-overhead, spatial-temporal computing due to the underlying motion compensation system's need for constant updates of a time-dependent point cloud model.

Several other works have explored deep learning methods for compensating for ego-motion during object detection and segmentation tasks. These approaches typically involve discretizing the event stream into a series of frames, which can then be used to train convolutional neural networks (CNNs) for computing the visual odometry of the scene. For example, Nitin et. al. \cite{sanket2020evdodgenet} creates a 3-channel event frame from an event stream where the first and second channels are the positive and negative event counts respectively and the third is the average time between events. This frame is then passed to a series of shallow neural networks which jointly compute the movements of the camera and observed objects. Zhu et. al \cite{zhu2019unsupervised} takes a similar approach though uses a novel method to discretize the time domain though uses bilinear sampling. The resultant frames are given to an encoder-decoder CNN which has another network tied to it's residual block for the purpose of pose estimation when performing ego-motion prediction.

Chen, et. al. \cite{DBLP:journals/corr/abs-1709-09323} introduces a non-biologically inspired method for performing standard computer vision tasks with ego-motion is presented. Their major contribution, as opposed to our work, is their evaluation of the effectiveness of their approach in sub-optimal viewing conditions such as those with motion blur and poor lighting. However, there is no mention of biological inspiration for this method. In contrast, our approach is inspired by biology and targets a broader range of object classes, including cars, pedestrians, buses, trucks, bicycles, riders, and motorcycles.

While the aforementioned studies were the closest to our research, there have been multiple other methods proposed and investigated in radar applications \cite{lee2020long} and frame-based vision \cite{li2022dynamic}.  Our approach is radically different from previous works because the underpinnings of object motion sensitivity come directly from biology rather than being defined by an arbitrary network architecture or parametric model.

\textbf{Size, Weight, and Power (SWaP)} State of the art applications of deep neural networks require quick and accurate processing of event-based sensory information that is compatible real-time intelligent systems.
Prior works in this area can be broken down into two major areas: algorithm optimization and in-sensor computation.
Algorithm optimization aims to process sensory information from existing technologies such as RGB or DVS and optimize the machine learning pipeline.
These models are responsible for extracting fundamental features from visual perceptions to make better decisions on lower-dimensional data.

The majority of literature focuses on the algorithmic optimization of computer vision systems to more standard RGB and DVS sensors.
A variety of sub-fields emerge with solutions that aim to mitigate the performance penalties that come with complicated neural network optimization systems such as quantization \cite{choi2016towards}, pruning \cite{blalock2020state}, neuromorphic computing \cite{schuman2022opportunities,patton2021neuromorphic}, and novel architecture design \cite{parsa2021multi, schuman2020automated}. While these works result in intelligent systems that are more capable in edge applications, a significant portion of the learning is dedicated to dealing with noisy and low entropy visual representations.

Numerous works strive to mitigate the challenges that hinder the wider adoption of intelligent computer vision systems in SWaP (size, weight, and power)-constrained environments by enhancing the sensor's processing capabilities via chip integration methodologies.
A variety of methods are explored with varying success such as 3D monolithic integration \cite{vivet2018monolithic}, 3D heterogeneous integration \cite{zhou2020near}, planar system on chip (SoC) integration \cite{zhou2020near}, and 2.5D chiplet integration \cite{zhou2020near}. These works emphasize on the hardware itself and propose it as a possible solution to a variety of problems. Our work differs in using biologically inspired OMS functionality for algorithmic analysis that can be embedded inside sensor arrays featuring spatio-temporal computations while leveraging 3D integration schemes.
 
In conclusion, there have been extensive research efforts dedicated to ego-motion detection and compensation, as well as dimensionality reduction for SWaP-constrained environments. While these areas have been investigated separately, there has been limited research at their intersection where the trade-off between ego-motion classification performance, information density, and applicability to existing computer vision applications is explored. This paper seeks to establish a research foundation at the intersection of ego-motion detection and compensation, while taking into account information density and applicability in SWaP-constrained environments through the use of a biologically inspired functionality - Object Motion Sensitivity.

\section{Research Methods}

\begin{figure}
    \centering
    \includegraphics[width=0.45\textwidth]{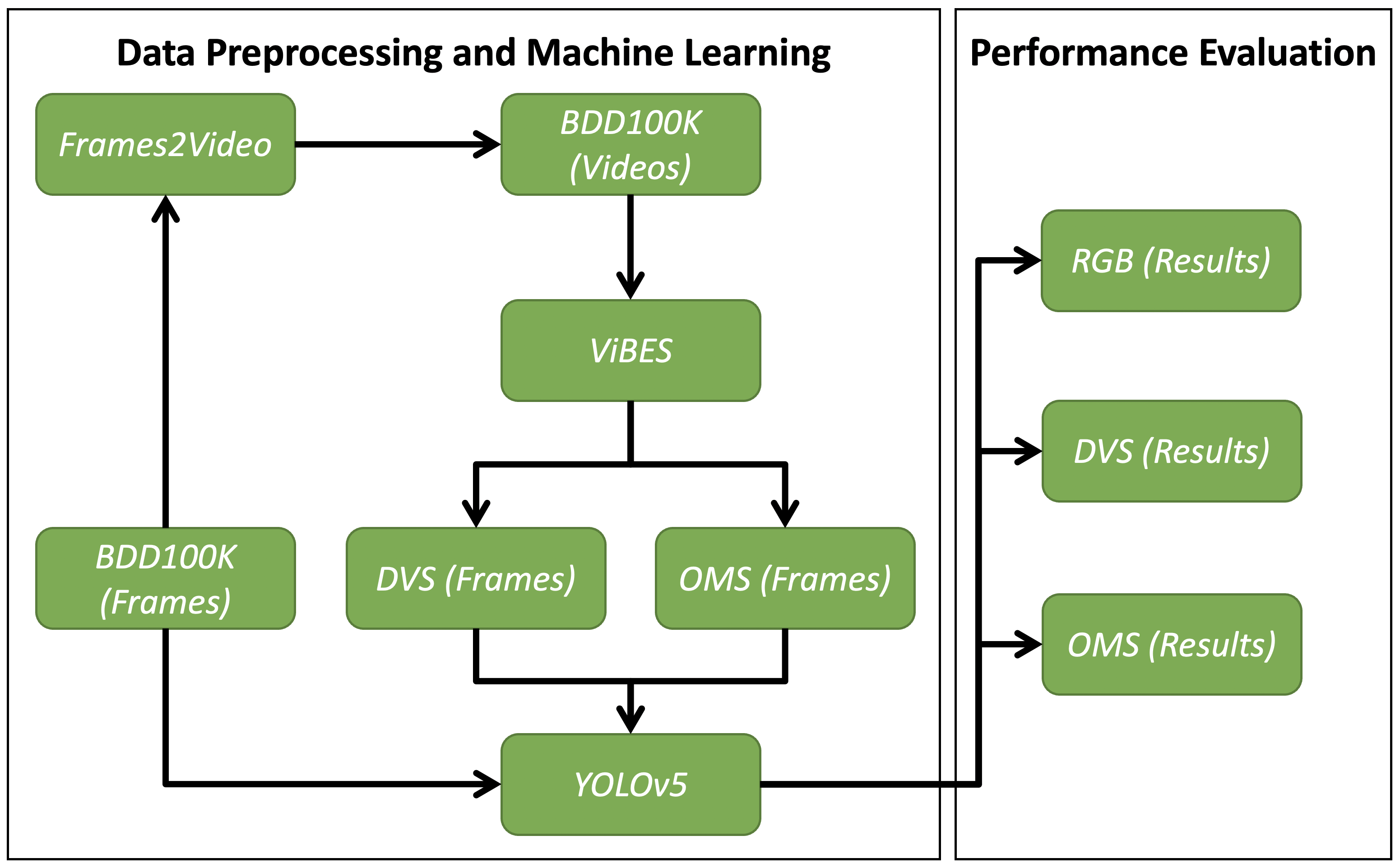}
    \caption{A flow chart presenting our processing for preprocessing BDD100k, converting it to DVS and OMS representations, fine-tuning YOLOv5, and our performance evaluation.}
    \label{fig:methods_flowchart}
\end{figure}

\noindent
In this section, we discuss the fundamental building blocks of this project. As shown in Figure \ref{fig:methods_flowchart}, we start off with a discussion about object motion sensitivity and Integrated Retinal Functionality in CMOS Image Sensors (IRIS). Next, we take a detailed look at the model architecture, machine learning task, and the various metrics used to evaluate the broader impacts and applications of Object Motion Sensitivity (OMS) on existing computer vision tasks.

\subsection{Dataset}
\noindent
Event-based cameras transmit a continuous stream of events corresponding to reflectance changes in the scene \cite{gallego2020event}. These events are transmitted in an address event representation (AER) unlike traditional active pixel sensors (APS) which output a 3-dimensional image. Due to this difference in output and the limited availability of event-based cameras, the creation of datasets has been limited to static cameras observing single moving objects, such as CIFAR10-DVS \cite{10.3389/fnins.2017.00309} and DHP19 \cite{9025364}. Recently, more complex datasets like DDD20 \cite{DBLP:journals/corr/abs-2005-08605} have been created using event cameras on a moving platform. However, they lack the accurate bounding box and segmentation labels required to compare the detection accuracy of a model trained on OMS data versus the detection accuracy of that same model trained on DVS data.

To overcome this limitation, we used the ViBES retinal computation simulator \cite{vibes} to construct frames that approximate DVS data from the Berkeley Deep Drive 100K Multi-Object Tracking and Segmentation (BDD100K MOTS) dataset \cite{DBLP:journals/corr/abs-1805-04687}.
The BDD100K MOTS dataset contains 90 videos from the original BDD100k dataset. The original videos were recorded at 30 fps and resampled to 5 fps for the MOTS dataset so that the salient objects in each frame could be labeled.
This results in each video having approximately 200 frames per video.
The observed objects fall into one of seven classes: cars, pedestrians, buses, trucks, bicycles, riders, and motorcycles.
We randomly chose 60 of these videos to convert into DVS data and then fed that DVS data into the algorithmic implementation of OMS provided by ViBES.
Our final dataset contains a close approximation of what one could obtain using an OMS inspired sensor such as IRIS \cite{yin2022iris}.

\subsection{Object Motion Sensitivity}

\begin{figure*}[ht!]
    \centering
    \includegraphics[width=\textwidth]{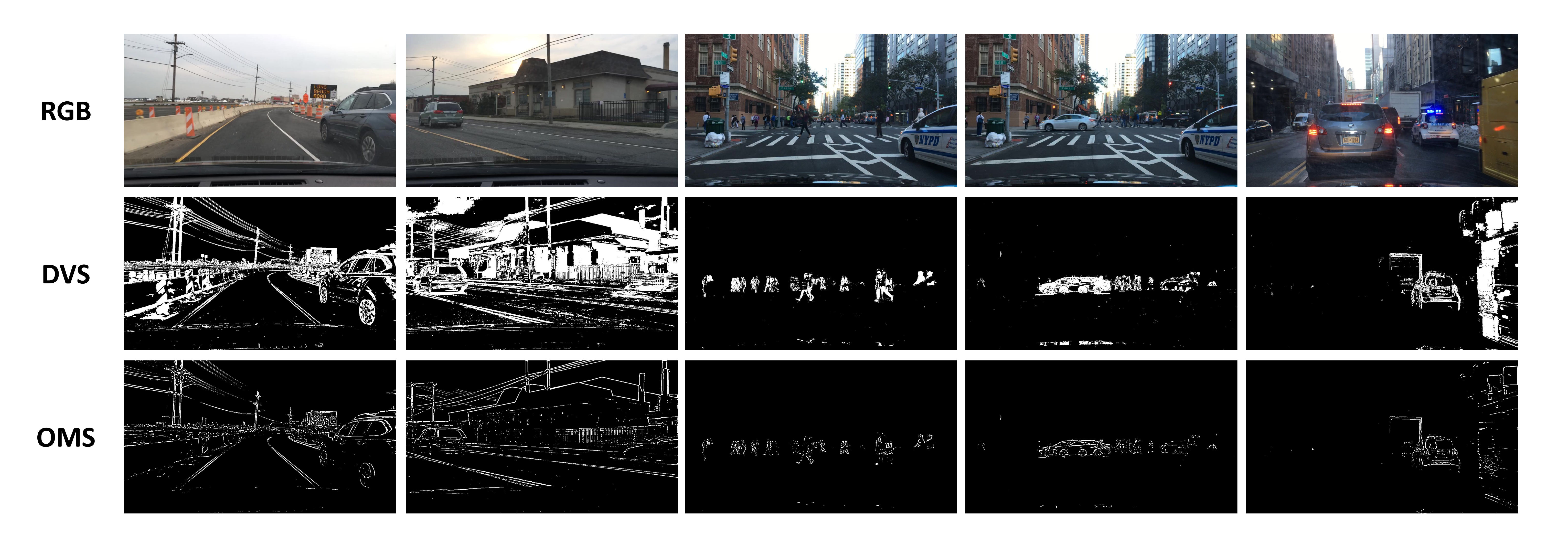}
    \caption{A visual comparison of the differences between RGB, DVS, and OMS. It is important to notice that OMS retains the majority of the spatial features of DVS while drastically reducing the noise and number of spikes.}
    \label{fig:image_comparison}
\end{figure*}

\begin{figure}
    \centering
    \includegraphics[width=0.48\textwidth]{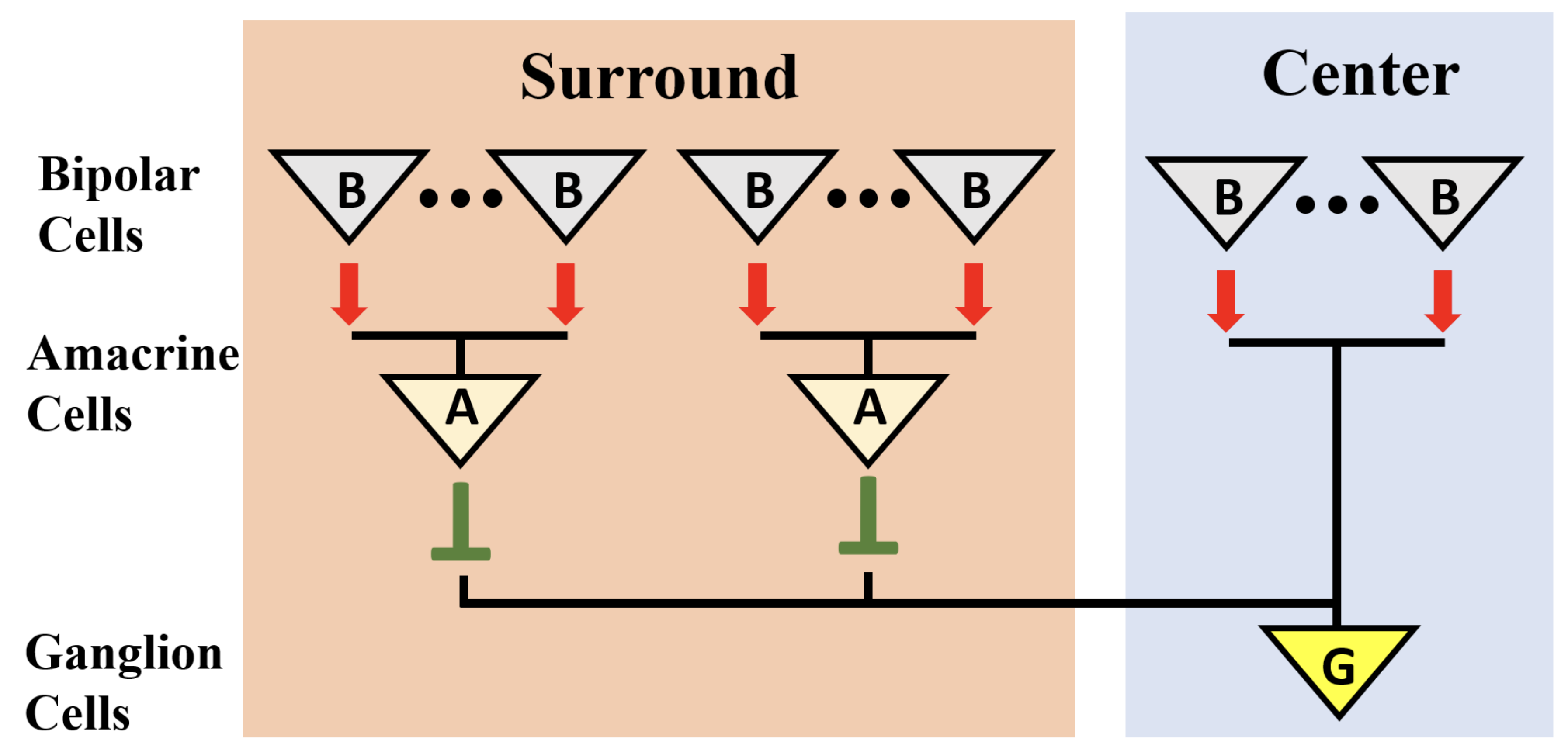}
    \caption{Retinal object motion sensitivity circuitry.}
    \label{fig:oms_circuit}
\end{figure}

\noindent
The visual pathways of many organisms have evolved the OMS circuit to suppress stimuli created by arbitrary eye movements and global motion. As shown in Figure \ref{fig:oms_circuit}, this circuit is comprised of bipolar, amacrine, and retinal ganglion cell layers, which work together to distinguish between stimuli created from global and local motion.
Figure \ref{fig:image_comparison} visually represents the similarity and differences between RGB, DVS, and OMS.
As shown in Algorithm \ref{oms}, The ViBES simulator contains an algorithmic approximation of the retinal OMS computation.
In the case of the BDD100k dataset, the data is initially in the form of a collection of RGB frames.
When executing on this data type, ViBES first computes frames approximating DVS by performing a difference on the frames and determining if the result exceeds the chosen contrast threshold of \(0.1\).

\begin{algorithm}[hbt!]
\caption{OMS algorithm}\label{oms}
\begin{algorithmic}[1]
\Require {radius $r_1$ $<$ radius $r_2$}
\Procedure{OMS}{$dvs\_frames$, $s\_weight$}
\State $center\gets diskfilter(r_1)$
\State $surround \gets s\_weight * diskfilter(r_2)$

\For{$i$ in $len(dvs\_frames)$}{}
    \State $frame \gets dvs\_frames[i]$
    \State $fil\_center \gets applyFilter(center, frame)$
    \State $fil\_surr \gets applyFilter(surround, frame)$
    \State $events \gets fil\_center - fil\_surr$
    \State $OMS[i] \gets events > threshold$
\EndFor

\State \textbf{return} $OMS$
\EndProcedure
\end{algorithmic}
\end{algorithm}

This method is an approximation of that performed by DVS cameras whose events defined as \(e_n=\{x_n, y_n, t_n, p_n\}\) are elicited when a change in the log photocurrent \(L=log(I)\) exceeds the temporal contrast threshold \(\pm C\) at a given pixel \(\varphi_n=\{x_n, y_n\}\) based upon \(E(\varphi_n, t_n)\) such that \cite{gallego2020event}:

\begin{equation}
\Delta L(\varphi_n, t_n)) = L(\varphi_n, t_n) - L(\varphi_{n-1}, t_{n-1})
\end{equation}

\begin{equation}
   E(\varphi_n, t_n)=\begin{cases}
     e_n, & \text{$\Delta L(\varphi_n, t_n) > +C)$} \\
     e_n, & \text{$\Delta L(\varphi_n, t_n) < -C)$} \\
     \varnothing, & otherwise
   \end{cases}
 \end{equation} \\

The resultant DVS frames are then sent to the OMS function \ref{oms}. From a biological perspective, the DVS frames represent the response to photoreceptor activation by bipolar cells, and the OMS function takes the role of the amacrine and retinal ganglion cell layers. The OMS algorithm is comprised of two circular averaging filters also known as disk filters \cite{MATLAB}. These filters are matrices containing a discrete feathered circle of a chosen radius whose values sum to one. Values in the center of the matrix possess larger values and thus carry more weight. The matrix convolves over the frame by centering itself over each pixel and storing resultant value into said pixel's position. This value is the mean contrast of the region covered by the disk filter.

The smaller of these disk filters is the center filter which represents a retinal ganglion cell (RGC) and the excitatory bipolar cell cluster with which it is connected. We chose a radius of 1 to lower the chance that a single cell cluster covers an entire entity. The larger disk filter serves the role of the amacrine cells, which are designed to inhibit the RGCs' response if global motion is observed. For this filter, a radius of 5 was chosen to cover a sufficiently wide region of each frame without significantly diminishing the weights. If the weights are too small then the surround filter will have no impact on the center filter values. In order to simulate the inhibition, the mean contrast values from the amacrine (surround) filter are subtracted from those of the RGC (center) filter. If the resultant values are larger than the threshold, we chose a threshold of 0.1 to match DVS, then a Boolean spike is stored in the OMS frame tensor.




\subsection{Integrated Retinal Functionality in Images Sensors (IRIS)}

\noindent
First proposed in \cite{yin2022iris}, IRIS cameras aim to embed retinal computations inside image sensing platforms, including object motion sensitivity (OMS). IRIS cameras are the next generation of neuromorphic cameras that aim to mimic feature-extraction computations within biological retina from photo-transduction to computations performed in the inner retinal layers, much of which has been recently discovered by the retinal neuroscience community \cite{schwartz2021object}. The initial version of the sensor implements two retinal features \cite{schwartz2021retinal}: object motion sensitivity and looming detection. 

In comparison to state-of-the-art active pixel CMOS image sensors (RGB cameras) \cite{feng2019computer, moeslund2001survey} that use a plethora of sophisticated and computationally complex algorithms to extract images features, retinal computations for IRIS cameras can be embedded inside an image sensor using low-cost, highly-efficient retina-inspired circuits \cite{shah2021review}. Similarly, IRIS cameras go beyond existing DVS cameras that focus on the changing luminance detection aspect of the retina to embed analog spatio-temporal computations of inner retinal layers needed for extracting retinal features by leveraging 3D integration of semiconductor chips \cite{choudhury20103d}. 

The outer retinal computations (bipolar cells functionality) can be implemented utilizing a modified active pixel sensor (APS) as well as the dynamic vision sensor (DVS) \cite{yin2022iris} on the back-side illuminated die. In contrast, the inner retinal circuits (e.g., amacrine and ganglion cells functionality of OMS features) can be implemented in a separate die and vertically stacked with the sensor die using pixel-parallel fine-pitched Cu-Cu hybrid bonding while maintaining the pixel density \cite{sony_3D}. With respect to the OMS algorithm, IRIS cameras distribute retinal computations from photo-transduction to RGCs using an interleaved center-surround receptive field distributed throughout the camera focal plane \cite{olveczky_Baccus_Meister_2007}.

The inner retinal circuits' excitatory (from the center receptive field) and inhibitory (from the surrounding receptive field) connections have been ensured by the opposite direction of current flows inside the CMOS circuits. A thresholding circuit compares the summed signal of the center and surrounding receptive field and generates an OMS feature spike when the summed signal crosses the OMS threshold. 
IRIS cameras thus form the required underlying hardware substrate that can implement OMS inside state-of-the-art camera manufacturing technology for real-time extraction of OMS spikes in highly SWaP (size, weight, and power) constrained environments.


\subsection{YOLOv5}

\noindent
Based on the initial version introduced in 2015 \cite{journals/corr/RedmonDGF15}, YOLOv5 (You Only Look Once Version 5) \cite{glenn_jocher_2022_7347926} stands as a major step forward as it introduces a variety of features over the initial version.
Throughout the different versions, multiple features have been added such as focal loss, batch normalization, and Mosaic data augmentation.
This model architecture has been deployed in a variety of scenarios such as autonomous vehicles \cite{kasper2021detecting}, medical imaging \cite{mohiyuddin2022breast}, and video surveillance \cite{wu2021application}. 
Accuracy for this model on object detection tasks is defined as the mean average precision (mAP). 
We chose YOLOv5 for this study due to its impressive performance and user-friendly interface, which facilitates the replication and extension of results.

\subsection{Bandwidth Reduction}

\noindent
To gain a quantitative perspective of the capabilities of OMS, we established a quantitative metric for bandwidth reduction represented by $bit\_rate$. 

\begin{equation}
    bit\_rate=hw\psi
    \label{bit_rate}
\end{equation}

\noindent
Our metric for bandwidth, as shown in Equation \ref{bit_rate}, is bit rate per frame which is defined as $bit\_rate=(hw\psi)$ where $h,w \in \mathbb{N} \geq 0$ and $\psi \in \mathbb{R} \geq 0$. $h$ and $w$ are the height and width of an individual frame in pixels. $\psi$ is the bit depth of an individual pixel. Our RGB images have a $\psi = 24$ with DVS and OMS representations having a $\psi = 1$.

A lower bit rate indicates a lower data bandwidth requirement while also reducing the total number of bits to be transmitted over a communication channel. Depending on the underlying communication scheme - wired \cite{kempainen2002low}, short-distance \cite{mozaffariahrar2022survey}, or long-distance wireless communication \cite{sinha2017survey}, the lower data bandwidth translates to lower communication energy while also avoiding data congestion in bandwidth-constrained environments.

\section{Results}

\noindent
We evaluated the effectiveness of OMS for an object detection task on the Berkeley Deep Drive dataset \cite{DBLP:journals/corr/abs-1805-04687}.
All machine learning tests were conducted with the small configuration of YOLOv5 which has 7.2 million parameters and requires 16.5 billion FLOPS.
This model was pretrained on the COCO dataset \cite{DBLP:journals/corr/LinMBHPRDZ14} for 300 epochs and results in a final mAP value of 37.4. The following subsections contain information and results from each of the performance comparisons used to evaluate the effectiveness of OMS versus RGB and DVS images.
We start off by evaluating the performance delta between RGB, DVS, and OMS in their native state with each image type having a resolution of 1280 by 720.
This resolution will remain static through all of the following tests.
For each image representation, we fine tune the aforementioned pretrained weights for 100 epochs with a batch size of 128 on a computer equipped with an Intel Xeon W-2295 processor, 128GB of system memory, and an Nvidia RTX A5000.
We continue with an evaluation of the average bit rate per frame, analogous to spike rate, of each image representation throughout the Berkeley dataset. We end this subsection by normalizing the performance values of each image type by their respective data rate per frame.

\subsection{Native Performance}

\noindent
We begin with a look at the native performance of each image type for object detection with YOLOv5.

\begin{figure}
    \centering
    \includegraphics[width=0.5\textwidth]{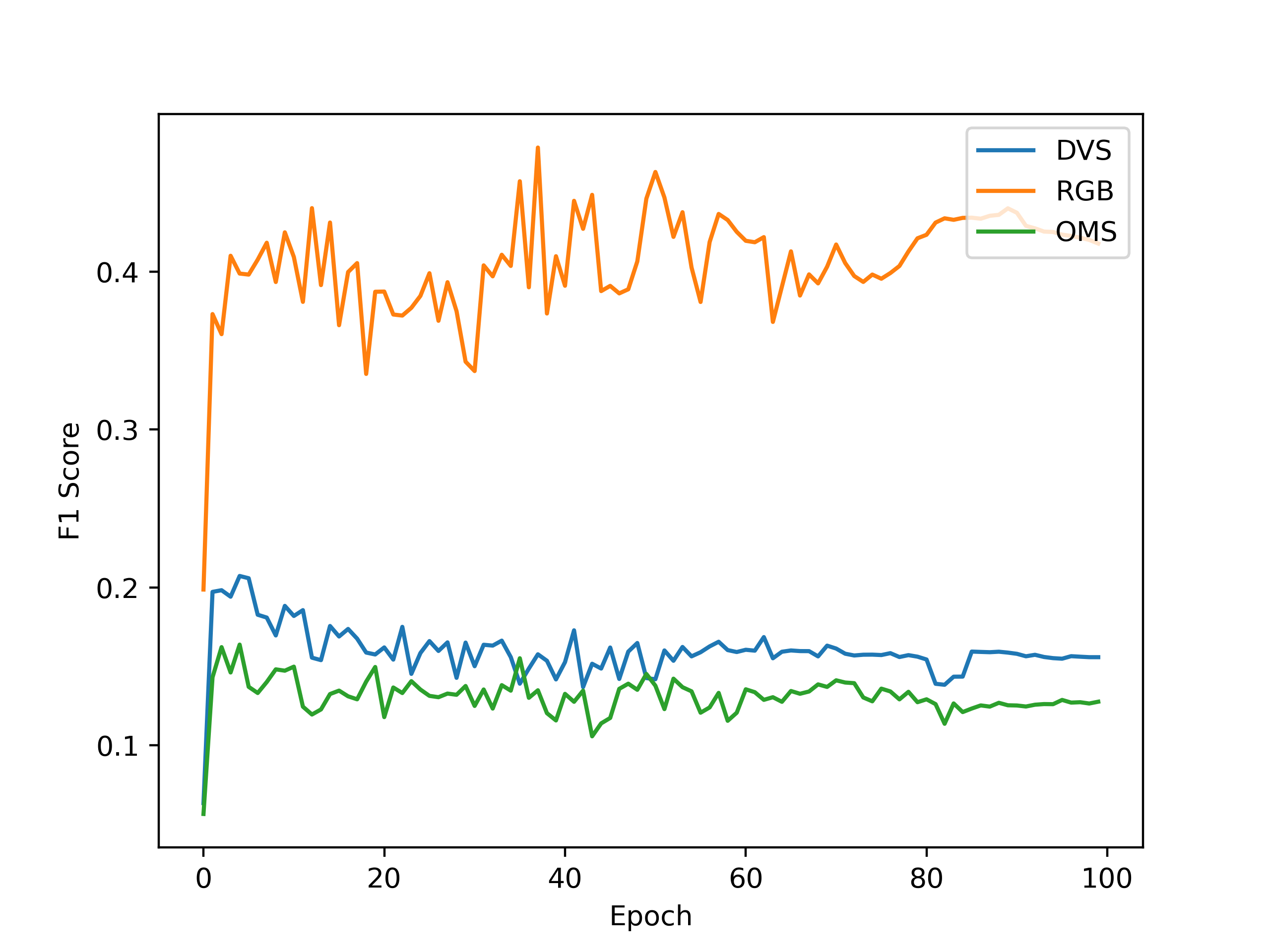}
    \caption{A line chart showing the performance deltas between the three image representations at every training epoch.}
    \label{fig:native_performance}
\end{figure}

As shown in Figure \ref{fig:native_performance}, RGB outperforms the other image types by a significant margin where DVS and OMS fall behind by $62.89$\% and $69.83$\%, respectively.
These performance penalties are to be expected as RGB image sensors capture the magnitude of different wavelengths of light at every pixel which results in a drastically higher data rate.
DVS' biologically inspired nature is designed to produce less information per frame.
This explanation is only amplified for the ever-increasing sparsity engineered into OMS sensors.
When applications are not limited by size, weight, and power constraints, it is evident that RGB is the best camera for this particular computer vision task.

\subsection{Data Rate}

\noindent
Given that OMS is designed to increase the feature-richness of individual spikes along with their sparsity, it is vital that we compare the average data rates of the individual representations across our dataset.

\begin{table}[h]
    \centering
    \begin{tabular}{|l|l|}
    \hline
    \textit{Type} & \textit{Bit Rate} \\ \hline
    RGB           & 2.21e7            \\
    DVS           & 1.96e5            \\
    OMS           & \textbf{3.77e4}   \\ \hline
    \end{tabular}
    \caption{A comparison of the average bit rate per frame of each image representation.}
    \label{table:data_rate}
\end{table}

Given the low entropy within RGB images, we expect it to have the highest data rate.
This comes in stark contrast to DVS and OMS where they are designed to increase information density while simultaneously decreasing data rates.
Table \ref{table:data_rate} shows the average data rates of these image representations across the entire BDD100K MOTS dataset.
As expected, we see that RGB has the highest data rate with a data rate of $2.21 \times 10^{7}$ bits per frame versus DVS and OMS with data rates of $1.96 \times 10^{5}$ bits per frame and $3.77 \times 10^{4}$ bits per frame, respectively.

\subsection{Performance versus Data Rate}

\noindent
With the drastic reductions in data rate among the more biologically inspired representations, we need to evaluate how much of an impact this will have on the overall F1-Score of our computer vision system. Therefore, we evaluate the performance of each image representation where the F1-Scores are normalized by the data rates.

\begin{figure}
    \centering
    \includegraphics[width=0.5\textwidth]{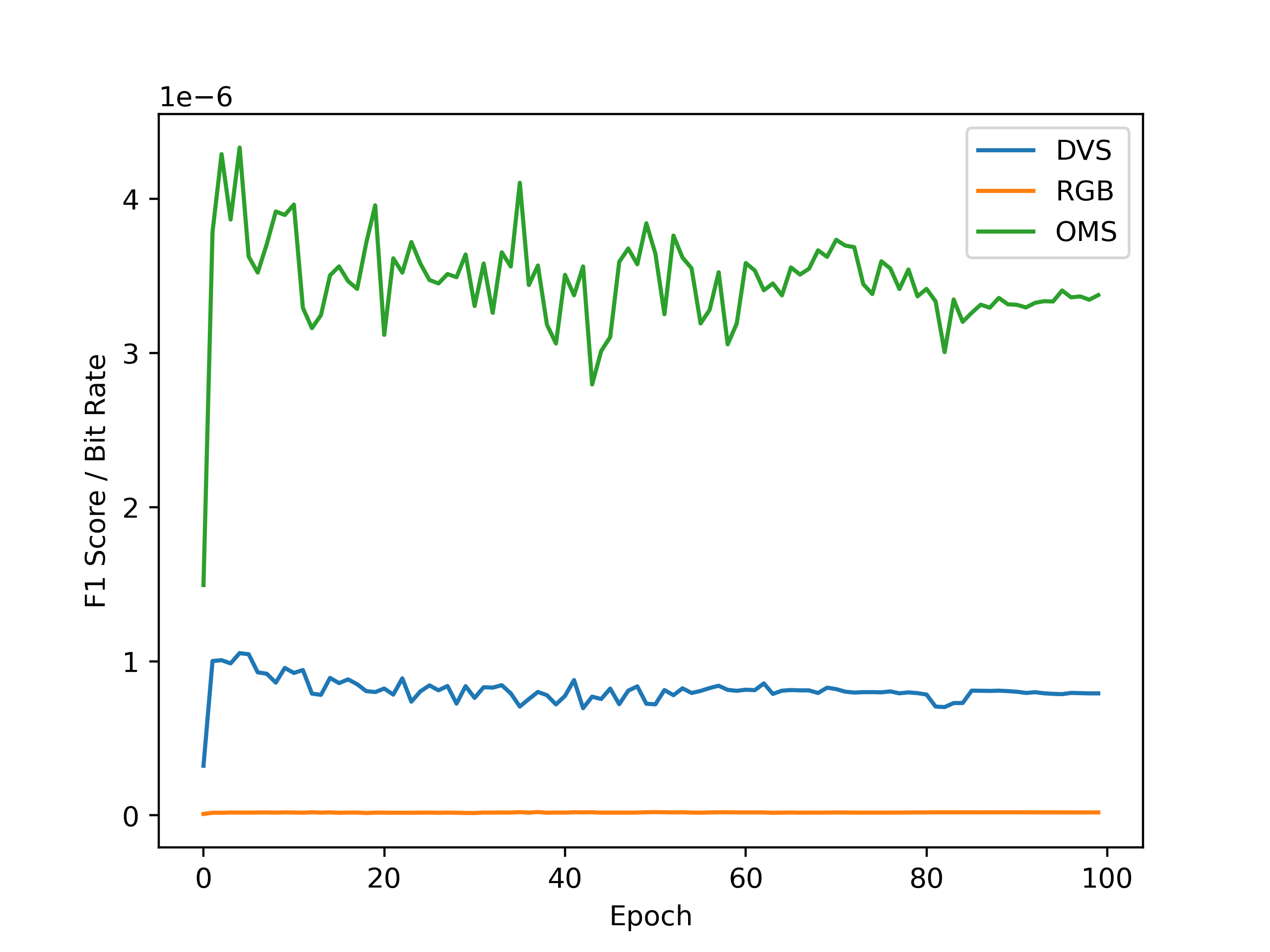}
    \caption{The F1 scores for the YOLOv5 models fine-tuned on RGB, DVS, and OMS where the final scores are normalized by the average bit rate per frame of the given representation.}
    \label{fig:performance_versus_data_rate}
\end{figure} 

Given the low information density yet high data rate of RGB we expect it to have the lowest coefficient of performance versus data rate. 
The more biologically inspired methods should present ratio increases from RGB to DVS and DVS to OMS.
Figure \ref{fig:performance_versus_data_rate} shows the performance versus data rate for RGB, DVS, and OMS.
The performance versus data rate values for each representation are $1.88 \times 10^{-8}$, $7.91 \times 10^{-7}$, and $3.37 \times 10^{-6}$, respectively.
These results highlight that, compared to RGB, individual DVS bits contribute $41.07$x more F1-Score per bit of information. OMS builds upon the impressive foundation of DVS with $187.25$x and $3.26$x more information per bit compared RGB and DVS, respectively.

\section{Discussion \& Future Works}

\noindent
In this work, we conducted a study of the applicability and effectiveness of object motion sensitivity (OMS) versus dynamic vision sensors (DVS), and traditional RGB-based sensors. OMS is a biological computation conducted within animal retinas where the goal is to reduce the dimensionality of visual information from the individual color values perceived at each cell to a more feature-rich and lower dimensional representation.

Rather than fully evaluate the biological plausibility of this mathematical representation of OMS, the focus of this paper was shifted to evaluate its application on more standard computer vision tasks. We choose object detection on the BDD100K MOTS dataset with our deep learning model being represented by YOLOv5. The mathematical representation for OMS is implemented within the Visual Behavioral Environment Simulator (ViBES). While this paper doesn't focus on the physical deployment of this algorithm, the hardware design has been developed and published in \cite{yin2022iris}.

We fine-tuned a pretrained version of YOLOv5 on both DVS and OMS for our performance evaluation on the BDD100K validation set.
We used three metrics to evaluate multiple performance attributes between the various image representations: F1-score, data rate, and F1-score versus data rate.

We began by showing that in their native state, without taking into consideration any of their unique performance characteristics, RGB is the most accurate representation with a final F1-score of $0.4177$.
The other types trailed by a significant margin with DVS and OMS having 62.89\% and 69.83\% less F1-Score, respectively.

This story becomes drastically more interesting when taking into consideration the sparsity and lower bandwidths of DVS and OMS versus RGB. For example, the RGB sensor has an average data rate of $2.21 \times 10^7$ bits per frame versus DVS and OMS with an average bits per frame of $1.96 \times 10^5$ and $3.77 \times 10^4$, respectively. When normalizing the F1-Scores from each representation by their respective bit rate per frame, we see that an individual bit of OMS contains orders of magnitude more information versus RGB and DVS with $178.25$x and $3.26$x more information per bit, respectively. In other words, this means that a given bit information within OMS contains $178.25$x more information than RGB images and $3.26$x more information than DVS.

Although we have demonstrated the promising information density and F1-Score versus data rate achieved by OMS compared to RGB and DVS, we have numerous future objectives and ambitions for this project:


\begin{enumerate}
    \item Investigate the impact on the overall effectiveness of OMS versus DVS and RGB when changing algorithmic hyper-parameters through a Bayesian optimization scheme
    \item Compare the operational characteristics of the OMS simulation algorithm against its biological counterpart to gain a more in-depth understanding of how representative it truly is
    \item Implement OMS on a proven DVS simulator such as v2e \cite{hu2021v2e} or ESIM \cite{Rebecq18corl}
    \item Create and incorporate other fundamental retinal computations within this framework to learn the trade-offs between them and how to incorporate multiple features into a holistic computer vision system
\end{enumerate}

\ifpeerreview \else
\section*{Acknowledgments}
The authors would like to thank...
\fi

\bibliographystyle{IEEEtran}
\bibliography{main}

\begin{thebibliography}{10}
\providecommand{\url}[1]{#1}
\csname url@samestyle\endcsname
\providecommand{\newblock}{\relax}
\providecommand{\bibinfo}[2]{#2}
\providecommand{\BIBentrySTDinterwordspacing}{\spaceskip=0pt\relax}
\providecommand{\BIBentryALTinterwordstretchfactor}{4}
\providecommand{\BIBentryALTinterwordspacing}{\spaceskip=\fontdimen2\font plus
\BIBentryALTinterwordstretchfactor\fontdimen3\font minus
  \fontdimen4\font\relax}
\providecommand{\BIBforeignlanguage}[2]{{%
\expandafter\ifx\csname l@#1\endcsname\relax
\typeout{** WARNING: IEEEtran.bst: No hyphenation pattern has been}%
\typeout{** loaded for the language `#1'. Using the pattern for}%
\typeout{** the default language instead.}%
\else
\language=\csname l@#1\endcsname
\fi
#2}}
\providecommand{\BIBdecl}{\relax}
\BIBdecl

\bibitem{Hayes:12}
\BIBentryALTinterwordspacing
T.~Hayes, ``Next-generation cell phone cameras,'' \emph{Opt. Photon. News},
  vol.~23, no.~2, pp. 16--21, Jan 2012. [Online]. Available:
  \url{https://www.optica-opn.org/abstract.cfm?URI=opn-23-2-16}
\BIBentrySTDinterwordspacing

\bibitem{ma20223d}
X.~Ma, W.~Ouyang, A.~Simonelli, and E.~Ricci, ``3d object detection from images
  for autonomous driving: a survey,'' \emph{arXiv preprint arXiv:2202.02980},
  2022.

\bibitem{snyder2021thor}
S.~E. Snyder and G.~Husari, ``Thor: A deep learning approach for face mask
  detection to prevent the covid-19 pandemic,'' in \emph{SoutheastCon
  2021}.\hskip 1em plus 0.5em minus 0.4em\relax IEEE, 2021, pp. 1--8.

\bibitem{zhou2022computer}
L.~Zhou, L.~Zhang, and N.~Konz, ``Computer vision techniques in
  manufacturing,'' \emph{IEEE Transactions on Systems, Man, and Cybernetics:
  Systems}, 2022.

\bibitem{gallego2020event}
G.~Gallego \emph{et~al.}, ``Event-based vision: A survey,'' \emph{IEEE
  transactions on pattern analysis and machine intelligence}, vol.~44, no.~1,
  pp. 154--180, 2020.

\bibitem{binas2017ddd17}
J.~Binas, D.~Neil, S.-C. Liu, and T.~Delbruck, ``Ddd17: End-to-end davis
  driving dataset,'' \emph{arXiv preprint arXiv:1711.01458}, 2017.

\bibitem{khan2017ego}
N.~H. Khan and A.~Adnan, ``Ego-motion estimation concepts, algorithms and
  challenges: an overview,'' \emph{Multimedia Tools and Applications}, vol.~76,
  pp. 16\,581--16\,603, 2017.

\bibitem{jia2021self}
S.~Jia, X.~Pei, X.~Jing, and D.~Yao, ``Self-supervised 3d reconstruction and
  ego-motion estimation via on-board monocular video,'' \emph{IEEE Transactions
  on Intelligent Transportation Systems}, vol.~23, no.~7, pp. 7557--7569, 2021.

\bibitem{kinsman2012ego}
T.~Kinsman \emph{et~al.}, ``Ego-motion compensation improves fixation detection
  in wearable eye tracking,'' in \emph{Proceedings of the Symposium on Eye
  Tracking Research and Applications}, 2012, pp. 221--224.

\bibitem{schwartz2021object}
G.~W. Schwartz and D.~Swygart, ``Object motion sensitivity,'' in \emph{Retinal
  Computation}.\hskip 1em plus 0.5em minus 0.4em\relax Elsevier, 2021, pp.
  230--244.

\bibitem{baccus2008retinal}
S.~A. Baccus, B.~P. {\"O}lveczky, M.~Manu, and M.~Meister, ``A retinal circuit
  that computes object motion,'' \emph{Journal of Neuroscience}, vol.~28,
  no.~27, pp. 6807--6817, 2008.

\bibitem{yin2022iris}
Z.~Yin, M.~A.-A. Kaiser, L.~O. Camara, M.~Camarena, M.~Parsa, A.~Jacob,
  G.~Schwartz, and A.~Jaiswal, ``Iris: Integrated retinal functionality in
  image sensors,'' \emph{bioRxiv}, pp. 2022--08, 2022.

\bibitem{Rep_Ryan_2022}
\BIBentryALTinterwordspacing
T.~D.-O.-. Rep.~Ryan, ``\BIBforeignlanguage{eng}{H.r.4346 - 117th congress
  (2021-2022): Chips and science act},'' Aug 2022. [Online]. Available:
  \url{http://www.congress.gov/}
\BIBentrySTDinterwordspacing

\bibitem{stoffregen2019event}
T.~Stoffregen \emph{et~al.}, ``Event-based motion segmentation by motion
  compensation,'' in \emph{Proceedings of the IEEE/CVF International Conference
  on Computer Vision}, 2019, pp. 7244--7253.

\bibitem{gallego2018unifying}
G.~Gallego \emph{et~al.}, ``A unifying contrast maximization framework for
  event cameras, with applications to motion, depth, and optical flow
  estimation,'' in \emph{Proceedings of the IEEE conference on computer vision
  and pattern recognition}, 2018, pp. 3867--3876.

\bibitem{liu2020globally}
D.~Liu, A.~Parra, and T.-J. Chin, ``Globally optimal contrast maximisation for
  event-based motion estimation,'' in \emph{Proceedings of the IEEE/CVF
  Conference on Computer Vision and Pattern Recognition}, 2020, pp. 6349--6358.

\bibitem{DBLP:journals/corr/abs-1803-04523}
\BIBentryALTinterwordspacing
A.~Mitrokhin, C.~Ferm{\"{u}}ller, C.~Parameshwara, and Y.~Aloimonos,
  ``Event-based moving object detection and tracking,'' \emph{CoRR}, vol.
  abs/1803.04523, 2018. [Online]. Available:
  \url{http://arxiv.org/abs/1803.04523}
\BIBentrySTDinterwordspacing

\bibitem{sanket2020evdodgenet}
N.~J. Sanket \emph{et~al.}, ``Evdodgenet: Deep dynamic obstacle dodging with
  event cameras,'' in \emph{2020 IEEE International Conference on Robotics and
  Automation (ICRA)}.\hskip 1em plus 0.5em minus 0.4em\relax IEEE, 2020, pp.
  10\,651--10\,657.

\bibitem{zhu2019unsupervised}
A.~Z. Zhu \emph{et~al.}, ``Unsupervised event-based learning of optical flow,
  depth, and egomotion,'' in \emph{Proceedings of the IEEE/CVF Conference on
  Computer Vision and Pattern Recognition}, 2019, pp. 989--997.

\bibitem{DBLP:journals/corr/abs-1709-09323}
\BIBentryALTinterwordspacing
N.~F.~Y. Chen, ``Pseudo-labels for supervised learning on event-based data,''
  \emph{CoRR}, vol. abs/1709.09323, 2017. [Online]. Available:
  \url{http://arxiv.org/abs/1709.09323}
\BIBentrySTDinterwordspacing

\bibitem{lee2020long}
J.~Lee \emph{et~al.}, ``Long-term displacement measurement of full-scale
  bridges using camera ego-motion compensation,'' \emph{Mechanical Systems and
  Signal Processing}, vol. 140, p. 106651, 2020.

\bibitem{li2022dynamic}
W.~Li \emph{et~al.}, ``Dynamic registration: Joint ego motion estimation and 3d
  moving object detection in dynamic environment,'' \emph{arXiv preprint
  arXiv:2204.12769}, 2022.

\bibitem{choi2016towards}
Y.~Choi, M.~El-Khamy, and J.~Lee, ``Towards the limit of network
  quantization,'' \emph{arXiv preprint arXiv:1612.01543}, 2016.

\bibitem{blalock2020state}
D.~Blalock \emph{et~al.}, ``What is the state of neural network pruning?''
  \emph{Proceedings of machine learning and systems}, vol.~2, pp. 129--146,
  2020.

\bibitem{schuman2022opportunities}
C.~D. Schuman, S.~R. Kulkarni, M.~Parsa, J.~P. Mitchell, P.~Date, and B.~Kay,
  ``Opportunities for neuromorphic computing algorithms and applications,''
  \emph{Nature Computational Science}, vol.~2, no.~1, pp. 10--19, 2022.

\bibitem{patton2021neuromorphic}
R.~Patton, C.~Schuman, S.~Kulkarni, M.~Parsa \emph{et~al.}, ``Neuromorphic
  computing for autonomous racing,'' in \emph{International Conference on
  Neuromorphic Systems 2021}, 2021, pp. 1--5.

\bibitem{parsa2021multi}
M.~Parsa \emph{et~al.}, ``Multi-objective hyperparameter optimization for
  spiking neural network neuroevolution,'' in \emph{2021 IEEE Congress on
  Evolutionary Computation (CEC)}.\hskip 1em plus 0.5em minus 0.4em\relax IEEE,
  2021, pp. 1225--1232.

\bibitem{schuman2020automated}
C.~D. Schuman, J.~P. Mitchell, M.~Parsa \emph{et~al.}, ``Automated design of
  neuromorphic networks for scientific applications at the edge,'' in
  \emph{2020 International Joint Conference on Neural Networks (IJCNN)}.\hskip
  1em plus 0.5em minus 0.4em\relax IEEE, 2020, pp. 1--7.

\bibitem{vivet2018monolithic}
P.~Vivet \emph{et~al.}, ``Monolithic 3d: An alternative to advanced cmos
  scaling, technology perspectives and associated design methodology
  challenges,'' in \emph{2018 25th IEEE International Conference on
  Electronics, Circuits and Systems (ICECS)}.\hskip 1em plus 0.5em minus
  0.4em\relax IEEE, 2018, pp. 157--160.

\bibitem{zhou2020near}
F.~Zhou and Y.~Chai, ``Near-sensor and in-sensor computing,'' \emph{Nature
  Electronics}, vol.~3, no.~11, pp. 664--671, 2020.

\bibitem{10.3389/fnins.2017.00309}
\BIBentryALTinterwordspacing
H.~Li, H.~Liu, X.~Ji, G.~Li, and L.~Shi, ``Cifar10-dvs: An event-stream dataset
  for object classification,'' \emph{Frontiers in Neuroscience}, vol.~11, 2017.
  [Online]. Available:
  \url{https://www.frontiersin.org/articles/10.3389/fnins.2017.00309}
\BIBentrySTDinterwordspacing

\bibitem{9025364}
E.~Calabrese \emph{et~al.}, ``Dhp19: Dynamic vision sensor 3d human pose
  dataset,'' in \emph{2019 IEEE/CVF Conference on Computer Vision and Pattern
  Recognition Workshops (CVPRW)}, 2019, pp. 1695--1704.

\bibitem{DBLP:journals/corr/abs-2005-08605}
\BIBentryALTinterwordspacing
Y.~Hu, J.~Binas, D.~Neil, S.~Liu, and T.~Delbr{\"{u}}ck, ``{DDD20} end-to-end
  event camera driving dataset: Fusing frames and events with deep learning for
  improved steering prediction,'' \emph{CoRR}, vol. abs/2005.08605, 2020.
  [Online]. Available: \url{https://arxiv.org/abs/2005.08605}
\BIBentrySTDinterwordspacing

\bibitem{vibes}
\BIBentryALTinterwordspacing
G.~Schwartz, ``Visual environment behavioral simulator,'' 2023. [Online].
  Available: \url{https://github.com/SchwartzNU/ViBES}
\BIBentrySTDinterwordspacing

\bibitem{DBLP:journals/corr/abs-1805-04687}
\BIBentryALTinterwordspacing
F.~Yu \emph{et~al.}, ``{BDD100K:} {A} diverse driving video database with
  scalable annotation tooling,'' \emph{CoRR}, vol. abs/1805.04687, 2018.
  [Online]. Available: \url{http://arxiv.org/abs/1805.04687}
\BIBentrySTDinterwordspacing

\bibitem{MATLAB}
\BIBentryALTinterwordspacing
T.~M. Inc., ``Matlab version: 9.13.0 (r2022b),'' Natick, Massachusetts, United
  States, 2022. [Online]. Available: \url{https://www.mathworks.com}
\BIBentrySTDinterwordspacing

\bibitem{schwartz2021retinal}
G.~Schwartz, \emph{Retinal Computation}.\hskip 1em plus 0.5em minus 0.4em\relax
  Academic Press, 2021.

\bibitem{feng2019computer}
X.~Feng, Y.~Jiang, X.~Yang, M.~Du, and X.~Li, ``Computer vision algorithms and
  hardware implementations: A survey,'' \emph{Integration}, vol.~69, pp.
  309--320, 2019.

\bibitem{moeslund2001survey}
T.~B. Moeslund and E.~Granum, ``A survey of computer vision-based human motion
  capture,'' \emph{Computer vision and image understanding}, vol.~81, no.~3,
  pp. 231--268, 2001.

\bibitem{shah2021review}
P.~Shah and S.~S. Rathod, ``Review of bio-inspired silicon retina: From cell to
  system level implementation,'' in \emph{2021 International Conference on
  Communication information and Computing Technology (ICCICT)}.\hskip 1em plus
  0.5em minus 0.4em\relax IEEE, 2021, pp. 1--13.

\bibitem{choudhury20103d}
D.~Choudhury, ``3d integration technologies for emerging microsystems,'' in
  \emph{2010 IEEE MTT-S international microwave symposium}.\hskip 1em plus
  0.5em minus 0.4em\relax IEEE, 2010, pp. 1--4.

\bibitem{sony_3D}
Y.~Kagawa, N.~Fujii, K.~Aoyagi, Y.~Kobayashi, S.~Nishi, N.~Todaka,
  S.~Takeshita, J.~Taura, H.~Takahashi, Y.~Nishimura \emph{et~al.}, ``Novel
  stacked cmos image sensor with advanced cu2cu hybrid bonding,'' in \emph{2016
  IEEE International Electron Devices Meeting (IEDM)}.\hskip 1em plus 0.5em
  minus 0.4em\relax IEEE, 2016, pp. 8--4.

\bibitem{olveczky_Baccus_Meister_2007}
B.~P. Ölveczky, S.~A. Baccus, and M.~Meister,
  ``\BIBforeignlanguage{en}{Retinal adaptation to object motion},''
  \emph{\BIBforeignlanguage{en}{Neuron}}, vol.~56, no.~4, p. 689–700, Nov
  2007.

\bibitem{journals/corr/RedmonDGF15}
\BIBentryALTinterwordspacing
J.~Redmon, S.~K. Divvala, R.~B. Girshick, and A.~Farhadi, ``You only look once:
  Unified, real-time object detection,'' \emph{CoRR}, vol. abs/1506.02640,
  2015. [Online]. Available: \url{http://arxiv.org/abs/1506.02640}
\BIBentrySTDinterwordspacing

\bibitem{glenn_jocher_2022_7347926}
\BIBentryALTinterwordspacing
G.~Jocher, A.~Chaurasia, A.~Stoken \emph{et~al.}, ``{ultralytics/yolov5: v7.0 -
  YOLOv5 SOTA Realtime Instance Segmentation},'' Nov. 2022. [Online].
  Available: \url{https://doi.org/10.5281/zenodo.7347926}
\BIBentrySTDinterwordspacing

\bibitem{kasper2021detecting}
Kasper-Eulaers, ``Detecting heavy goods vehicles in rest areas in winter
  conditions using yolov5,'' \emph{Algorithms}, vol.~14, no.~4, p. 114, 2021.

\bibitem{mohiyuddin2022breast}
A.~Mohiyuddin \emph{et~al.}, ``Breast tumor detection and classification in
  mammogram images using modified yolov5 network,'' \emph{Computational and
  Mathematical Methods in Medicine}, vol. 2022, pp. 1--16, 2022.

\bibitem{wu2021application}
W.~Wu \emph{et~al.}, ``Application of local fully convolutional neural network
  combined with yolo v5 algorithm in small target detection of remote sensing
  image,'' \emph{PloS one}, vol.~16, no.~10, p. e0259283, 2021.

\bibitem{kempainen2002low}
S.~Kempainen, ``Low-voltage differential signaling (lvds),'' \emph{Altera
  Co-operation}, 2002.

\bibitem{mozaffariahrar2022survey}
E.~Mozaffariahrar, F.~Theoleyre, and M.~Menth, ``A survey of wi-fi 6:
  Technologies, advances, and challenges,'' \emph{Future Internet}, vol.~14,
  no.~10, p. 293, 2022.

\bibitem{sinha2017survey}
R.~S. Sinha \emph{et~al.}, ``A survey on lpwa technology: Lora and nb-iot,''
  \emph{Ict Express}, vol.~3, no.~1, pp. 14--21, 2017.

\bibitem{DBLP:journals/corr/LinMBHPRDZ14}
\BIBentryALTinterwordspacing
T.~Lin, M.~Maire \emph{et~al.}, ``Microsoft {COCO:} common objects in
  context,'' \emph{CoRR}, vol. abs/1405.0312, 2014. [Online]. Available:
  \url{http://arxiv.org/abs/1405.0312}
\BIBentrySTDinterwordspacing

\bibitem{hu2021v2e}
Y.~Hu, S.-C. Liu, and T.~Delbruck, ``v2e: From video frames to realistic dvs
  events,'' in \emph{Proceedings of the IEEE/CVF Conference on Computer Vision
  and Pattern Recognition}, 2021, pp. 1312--1321.

\bibitem{Rebecq18corl}
H.~Rebecq, D.~Gehrig, and D.~Scaramuzza, ``{ESIM}: an open event camera
  simulator,'' \emph{Conf. on Robotics Learning (CoRL)}, Oct. 2018.

\end{thebibliography}

\ifpeerreview \else


\begin{IEEEbiography}{Michael Shell}
Biography text here.
\end{IEEEbiography}


\begin{IEEEbiographynophoto}{John Doe}
Biography text here.
\end{IEEEbiographynophoto}


\fi

\end{document}